\documentclass[runningheads]{llncs}

\usepackage[T1]{fontenc}

\usepackage{graphicx}
\usepackage{comment}
\usepackage{amsmath,amssymb}
\usepackage{color}
\usepackage{url}
\usepackage{hyperref}

\usepackage{booktabs}
\usepackage{xcolor}
\usepackage{bookmark}
\usepackage{algorithm,algpseudocode}
\usepackage{subcaption}
\usepackage{appendix}

\newif\ifreview
\reviewfalse

\ifreview
	\usepackage{lineno}

	\linenumbers
\fi

\begin{document}


\def\SubNumber{094}

\def\GCPRTrack{Fast Review Track}

\def\MethodName{LADB}
\newcommand{\iali}[1]{\begin{align}#1\end{align}}
\newcommand{\ialid}[1]{\begin{aligned}#1\end{aligned}}
\newcommand{\ieqn}[1]{\begin{equation}#1\end{equation}}
\newcommand{\ieqa}[1]{\begin{eqnarray}#1\end{eqnarray}}
\newcommand{\imtx}[1]{\begin{bmatrix}#1\end{bmatrix}}

\newcommand{\be}{\begin{enumerate}}
\newcommand{\ee}{\end{enumerate}}
\newcommand{\bi}{\begin{itemize}}
\newcommand{\ei}{\end{itemize}}
\newcommand{\ba}{\begin{array}}
\newcommand{\ea}{\end{array}}
\newcommand{\bt}{\begin{tabular}}
\newcommand{\et}{\end{tabular}}
\newcommand{\btb}{\begin{tabbing}}
\newcommand{\etb}{\end{tabbing}}
\newcommand{\bfg}{\begin{figure}}
\newcommand{\efg}{\end{figure}}
\newcommand{\bsl}{\begin{slide}}
\newcommand{\esl}{\end{slide}}
\newcommand{\bthm}{\begin{theorem}}
\newcommand{\ethm}{\end{theorem}}
\newcommand{\bcor}{\begin{corollary}}
\newcommand{\ecor}{\end{corollary}}
\newcommand{\blem}{\begin{lemma}}
\newcommand{\elem}{\end{lemma}}
\newcommand{\bprop}{\begin{proposition}}
\newcommand{\eprop}{\end{proposition}}
\newcommand{\basm}{\begin{assumption}}
\newcommand{\easm}{\end{assumption}}
\newcommand{\baxm}{\begin{axiom}}
\newcommand{\eaxm}{\end{axiom}}
\newcommand{\bdfn}{\begin{definition}}
\newcommand{\edfn}{\end{definition}}
\newcommand{\brmk}{\begin{remark}}
\newcommand{\ermk}{\end{remark}}
\newcommand{\balg}{\begin{algorithm}}
\newcommand{\ealg}{\end{algorithm}}
\newcommand{\bntn}{\begin{notation}}
\newcommand{\entn}{\end{notation}}
\newcommand{\bexm}{\begin{example}}
\newcommand{\eexm}{\end{example}}
\newcommand{\bpf}{\begin{proof}}
\newcommand{\epf}{\end{proof}}

\newcommand{\tr}{\textcolor{red}}
\newcommand{\tb}{\textcolor{blue}}
\newcommand{\eps}{\varepsilon}
\newcommand{\0}{\mathbf{0}}
\newcommand{\1}{\mathbf{1}}
\newcommand{\st}{\mathrm{s.t.~}}
\newcommand{\iaoi}{{\rm if~and~only~if~}}
\newcommand{\ift}{\raisebox{-1.0ex}{$\check{}$}}
\newcommand{\bR}{\mathbb{R}}
\newcommand{\exR}{\overline{\mathbb{R}}}
\newcommand{\bC}{\mathbb{C}}
\newcommand{\bE}{\mathbb{E}}
\newcommand{\bQ}{\mathbb{Q}}
\newcommand{\bZ}{\mathbb{Z}}
\newcommand{\bN}{\mathbb{N}}
\newcommand{\bB}{\mathbb{B}}
\newcommand{\bM}{\mathbb{M}}
\newcommand{\bP}{\mathbb{P}}
\newcommand{\bV}{\mathbb{V}}
\newcommand{\bS}{\mathbb{S}}
\newcommand{\bH}{\mathbb{H}}
\newcommand{\cL}{\mathcal{L}}
\newcommand{\cF}{\mathcal{F}}
\newcommand{\cN}{\mathcal{N}}
\newcommand{\cK}{\mathcal{K}}
\newcommand{\cD}{\mathcal{D}}
\newcommand{\cG}{\mathcal{G}}
\newcommand{\cR}{\mathcal{R}}
\newcommand{\cE}{\mathcal{E}}
\newcommand{\cT}{\mathcal{T}}
\newcommand{\cC}{\mathcal{C}}
\newcommand{\cS}{\mathcal{S}}
\newcommand{\cA}{\mathcal{A}}
\newcommand{\cB}{\mathcal{B}}
\newcommand{\cI}{\mathcal{I}}
\newcommand{\cM}{\mathcal{M}}
\newcommand{\cP}{\mathcal{P}}
\newcommand{\cQ}{\mathcal{Q}}
\newcommand{\cH}{\mathcal{H}}
\newcommand{\cV}{\mathcal{V}}
\newcommand{\cU}{\mathcal{U}}
\newcommand{\cO}{\mathcal{O}}
\newcommand{\cX}{\mathcal{X}}
\newcommand{\cY}{\mathcal{Y}}
\newcommand{\cZ}{\mathcal{Z}}
\newcommand{\cW}{\mathcal{W}}
\newcommand{\fA}{\mathbf{A}}
\newcommand{\fD}{\mathbf{D}}
\newcommand{\fJ}{\mathbf{J}}
\newcommand{\fI}{\mathbf{I}}
\newcommand{\fB}{\mathbf{B}}
\newcommand{\fK}{\mathbf{K}}
\newcommand{\fX}{\mathbf{X}}
\newcommand{\fW}{\mathbf{W}}
\newcommand{\fR}{\mathbf{R}}
\newcommand{\fY}{\mathbf{Y}}
\newcommand{\fT}{\mathbf{T}}
\newcommand{\fu}{\mathbf{u}}
\newcommand{\fb}{\mathbf{b}}
\newcommand{\fv}{\mathbf{v}}
\newcommand{\fp}{\mathbf{p}}
\newcommand{\fq}{\mathbf{q}}
\newcommand{\fx}{\mathbf{x}}
\newcommand{\fg}{\mathbf{g}}
\newcommand{\fl}{\mathbf{l}}
\newcommand{\fe}{\mathbf{e}}
\newcommand{\fz}{\mathbf{z}}
\newcommand{\ft}{\mathbf{t}}
\newcommand{\fc}{\mathbf{c}}
\newcommand{\fn}{\mathbf{n}}
\newcommand{\fy}{\mathbf{y}}
\newcommand{\fd}{\mathbf{d}}
\newcommand{\fm}{\mathbf{m}}
\newcommand{\kX}{\mathfrak{X}}
\newcommand{\kF}{\mathfrak{F}}
\newcommand{\vp}{\vec{p}}
\newcommand{\ve}{\vec{e}}
\renewcommand{\hat}{\widehat}
\renewcommand{\tilde}{\widetilde}
\newcommand{\norm}[1]{\left\|#1\right\|}
\newcommand{\ip}[2]{\left\langle#1,#2\right\rangle}
\newcommand{\kld}[2]{\KL\left[#1\,\middle\|\,#2\right]}
\newcommand{\xpt}[2]{\mathbb{E}_{#1}\left[#2\right]}
\newcommand{\odeslv}{\mathrm{ODESolve}}
\newcommand{\ddim}{\mathrm{DDIM}}

\title{\MethodName: Latent Aligned Diffusion Bridges for Semi-Supervised Domain Translation}

\ifreview
\else
	\titlerunning{Latent Aligned Diffusion Bridges}

	\author{Xuqin Wang\inst{1,2} \and
	Tao Wu\inst{1} \and
	Yanfeng Zhang\inst{1} \and
	Lu Liu\inst{1} \and
	Dong Wang\inst{1} \and
	Mingwei Sun\inst{1} \and
	Yongliang Wang\inst{1} \and
	Niclas Zeller\inst{3} \and
	Daniel Cremers\inst{2}}
	
	\authorrunning{X. Wang, T. Wu et al.}
	
	\institute{ 
	Huawei Technologies \\
	\and Technical University of Munich \\
	\and Karlsruhe University of Applied Sciences \\
	  }
\fi

\maketitle              

\graphicspath{{./assets/}}

\begin{abstract}
Diffusion models excel at generating high-quality outputs but face challenges in data-scarce domains, where exhaustive retraining or costly paired data are often required. To address these limitations, we propose Latent Aligned Diffusion Bridges (LADB), a semi-supervised framework for sample-to-sample translation that effectively bridges domain gaps using partially paired data. By aligning source and target distributions within a shared latent space, LADB seamlessly integrates pretrained source-domain diffusion models with a target-domain Latent Aligned Diffusion Model (LADM), trained on partially paired latent representations. This approach enables deterministic domain mapping without the need for full supervision.
Compared to unpaired methods, which often lack controllability, and fully paired approaches that require large, domain-specific datasets, LADB strikes a balance between fidelity and diversity by leveraging a mixture of paired and unpaired latent-target couplings. Our experimental results demonstrate superior performance in depth-to-image translation under partial supervision. Furthermore, we extend LADB to handle multi-source translation (from depth maps and segmentation masks) and multi-target translation in a class-conditioned style transfer task, showcasing its versatility in handling diverse and heterogeneous use cases.
Ultimately, we present LADB as a scalable and versatile solution for real-world domain translation, particularly in scenarios where data annotation is costly or incomplete.
\end{abstract}

\section{Introduction}
\label{sec:intro}
Diffusion models excel at modeling complex data distributions by gradually denoising random noises into structured outputs. They have achieved remarkable successes in image generation, producing outputs with high perceptual quality \cite{song2021score,rombach2022high,karras2022elucidating}.
Moreover, various works have advanced their capabilities for controllable generation, incorporating text, masks, geometric priors, and other conditions \cite{rombach2022high,zhao2023uni}. 
Research has also expanded into more complex domains, such as novel view synthesis and 3D generation \cite{elata2024novelviewsynthesispixelspace,gupta20233dgentriplanelatentdiffusion,nichol2022pointegenerating3dpoint}.

While diffusion models excel in domains with abundant data, their performance degrades in data-scarce settings. 
In 3D generation for instance, photometric data collection requires dense multi-view data which are labor-intensive to collect and process \cite{lee2024semcitysemanticscenegeneration}. 
Other approaches rely on in-domain expertise to create assets \cite{fu20213dfront3dfurnishedrooms,wu2024blockfusionexpandable3dscene}.
In addtion, training of 3D diffusion models in \cite{zhai2023commonscenesgeneratingcommonsense3d,tang2024diffuscene} requires human annotations and expert-crafted rules for precise scene graph construction.

\begin{figure}[t]
  \centering
  \includegraphics[width=0.8\linewidth]{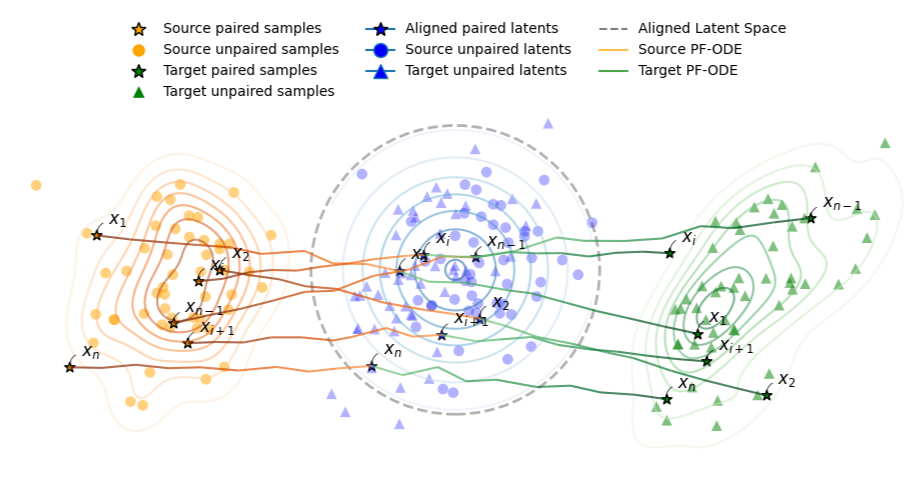}
  \vspace{-.3cm}
  \caption{\textbf{Latent Aligned Diffusion Bridges}.
  LADB achieves a cycle-consistent translation from \textcolor{orange}{source} to \textcolor{teal}{target} domain with partially available source-target couplings (marked by {\footnotesize$\bigstar$}), through a semi-supervised learning framework. The source and target samples are aligned in a common \textcolor{teal}{latent} space, supervised by (limited) source-to-target annotations. An LADB trajectory is formed by concatenation of the source and target PF-ODEs, thus preserving cycle consistency in the training data.
}
  \label{fig:LADM_triangle}
  \vspace{-.5cm}
\end{figure}

In this context, transferring knowledge from one or multiple source domains to a target domain has strong potentials to leverage the strengths of existing generative models for new scenarios. 
Ideally, any sample from known domains could be translated to the target domain with the assistance of generative models. 
However, major challenges persist. Firstly, aligning distributions across domains is non-trivial \cite{radford2021learningtransferablevisualmodels,zhu2017unpaired,choi2018starganunifiedgenerativeadversarial}. 
Also, translation must retain structural consistency to ensure meaningful cross-domain mappings, yet existing diffusion-based methods struggle to preserve fine-grained details across diverse scenarios \cite{wang2022zeroshotimagerestorationusing,meng2022sdeditguidedimagesynthesis,saharia2022paletteimagetoimagediffusionmodels}.
Besides, generalization across domains is tedious and often requires fully paired data or domain-specific tuning \cite{zhang2023adding,liu2023issimagesteppingstone}.

Sample-to-sample translation emerges as a promising solution for generative model-based translation to new domains. Since generative models are designed to sample from fitted data distributions, translation naturally occurs if the sample distributions are aligned. 
However, existing approaches face trade-offs. 
Unpaired translation methods \cite{zhu2017unpaired,su2023dual,de2024schrodinge} require no supervision but lack controllability in translation. 
Fully paired translation methods, such as bridge models \cite{liu2023i2sb,li2023bbdm,zhou2024denoising}, define bidirectional diffusion processes between domains but require per-domain training.
Conditional diffusion models \cite{rombach2022high,zhang2023adding,zhao2023uni} incorporate various priors for translation but necessitate architectural changes for new conditions, thus limiting their extensibility.
Crucially, none of these works commits to efficient utilization of partially paired data to balance controllability and generality.

To this end, we propose Latent Aligned Diffusion Bridges (LADB), a semi-supervised learning framework that leverages partially paired data for content-consistent domain translation; see Fig.~\ref{fig:LADM_triangle} for a schematic overview. 
More concretely, we utilize a pretrained source-domain latent diffusion model to translate limited source-to-target paired correspondences deterministically into latent-to-target alignments. This is followed by semi-supervised learning of a latent aligned diffusion model (LADM) from a mixture of paired and unpaired latent-target couples. The concatenation of source and target diffusions forms a Latent Aligned Diffusion Bridge (LADB) for sample-to-sample translation during inference. This framework is seamlessly extensible to multi-source and multi-target settings, opening up possibilities for a wide scope of downstream tasks.

\begin{figure*}[t]
  \centering
   \includegraphics[width=\linewidth]{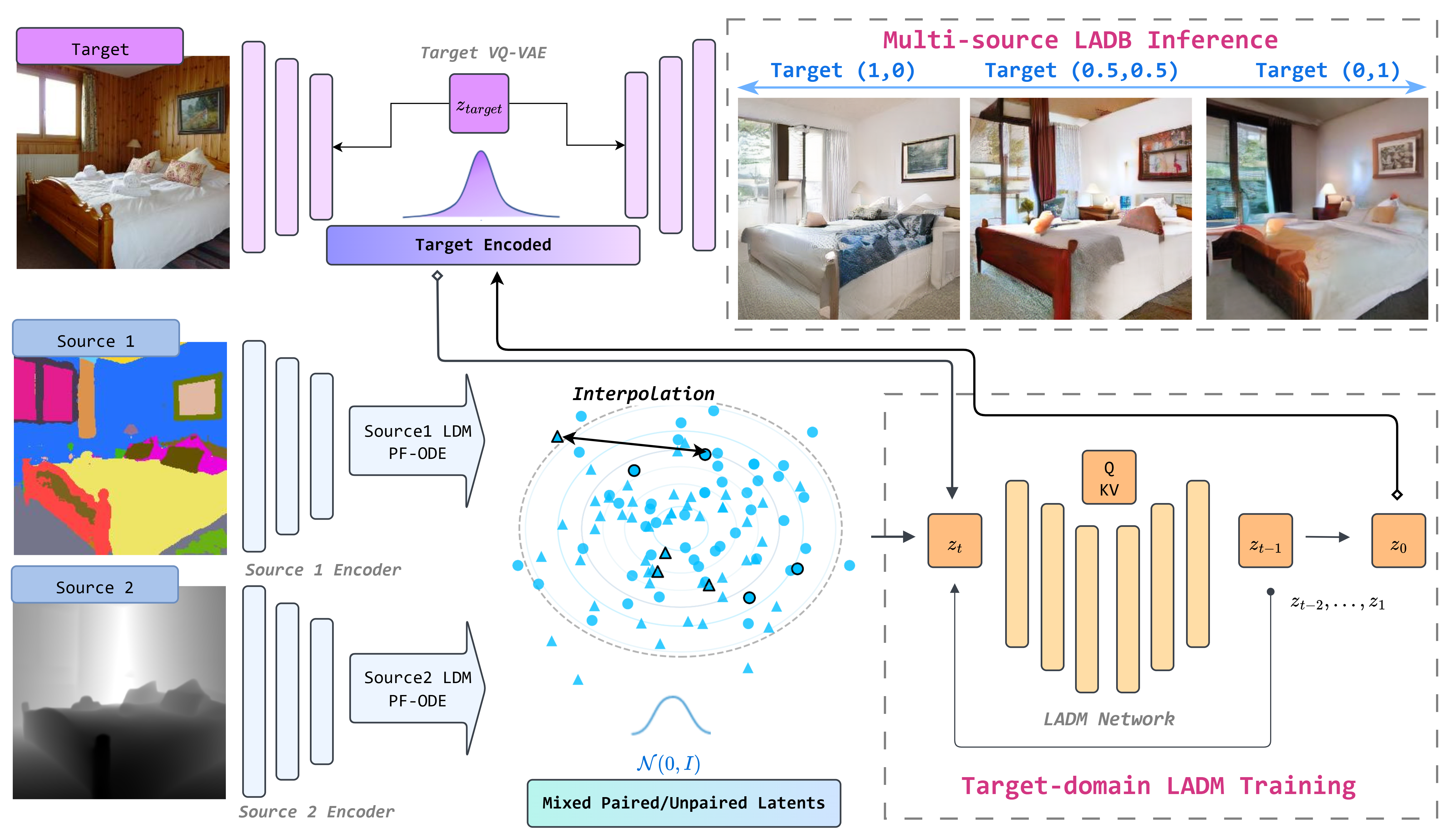}
   \caption{
  Model architecture (training and inference).
  \textbf{LADM training}:
(1) Infer latents from source(s) using pretrained source LDM(s), then construct paired latent-to-target correspondences.
(2) Couple paired and unpaired (random) latents with target samples into a mixture distribution. 
(3) Train target-domain LADM with the mixutre distribution via score matching.
   \textbf{LADB inference}:
  (1) Source-to-latent translation with pretrained source LDM. In case of multi-source inputs, addtionally perform interpolation on translated latents.
  (2) Latent-to-target translation with trained target-domain LADM.
  }
  \label{fig:architecture}
  \vspace{-.5cm}
\end{figure*}

In the experiments, we first present our LADB on depth-to-image translation compared favorably against other approaches on a suite of metrics and show our superiority under partial supervision.
Our experiments continue with image translation from cross-modal sources (depth maps and semantic segmentation masks), highlighting smooth source interpolation achieved by LADB.

\section{Preliminaries}
\label{sec:backgd}

We recap prior works on score-based diffusion and bridge models, based on which our Latent Aligned Diffusion Bridge (LADB) is proposed.

\subsection{Score-based Diffusion Models}
Given a $d$-dimensional data distribution $q_0(x_0)$, the diffusion model \cite{ho2020denoising,song2021score}, initiated with a sample $x_0\sim q_0$, follows a stochastic differential equation (SDE):
\iali{
    dx_t = f(x_t,t)dt + g(t)dw_t, \label{eq:f-sde}
}
with $t\in[0,1]$ the time variable, $f:\bR^d\times[0,1]\to\bR^d$ the vector-valued drift, $g:[0,1]\to\bR_{+}$ the scalar-valued diffusion coefficient, and $(w_t)_{0\leq t\leq 1}$ the standard Wiener process in $\bR^d$. The terminal distribution $q_1(x_1)$ at $t=1$ is an easy-to-sample prior distribution, e.g., a standard Gaussian. 

To generate data samples from a noise $x_1\sim q_1$, one simulates 
a probability flow ordinary differential equation (PF-ODE) \cite{song2021denoising} from $t=1$ to $0$:
\iali{
    dx_t &= \Big[f(x_t,t)-\frac12g^2(t)\nabla_{x_t}\log q_t(x_t)\Big]dt, \label{eq:pf-ode}
}
with the score function $\nabla_{x_t}\log q_t(x_t)$ of the time-dependent marginal $q_t$. 
The reverse-time PF-ODE \eqref{eq:pf-ode} preserves the marginals $(q_t)_{0\leq t\leq 1}$ from \eqref{eq:f-sde}. 

The unknown score function is typically parameterized as a neural network $s_\theta$ and trained with the denoising score matching loss \cite{vincent2011connection}:
\iali{
    \min_\theta & \bE_{\substack{t\sim\cU(0,1),\,x_0\sim q_0,\\x_1\sim q_1,\,x_t=\alpha_tx_0 + \sigma_tx_1}} 
    \Big[ \omega(t)\big\|s_\theta(x_t,t) - \nabla_{x_t}\log q(x_t|x_0)\big\|^2 \Big]. \label{eq:sml}
}
The transition kernel admits an analytical form such as
$q(x_t|x_0) = \cN(\alpha_tx_0, \sigma_t^2I)$,
which renders the score matching loss in \eqref{eq:sml} tractable. 

\subsection{Dual Diffusion Implicit Bridges}
The PF-ODE \eqref{eq:pf-ode}, with learned score $s_\theta$, enables uniquely identifiable encoding \cite{song2021score} through the drift:
\iali{
    v[s_\theta](x_t,t)=f(x_t,t)-\frac12g^2(t)s_\theta(x_t,t),
}
via the first-order integrator:
\iali{
    \odeslv(x(t_0), s_\theta, t_0, t_1) = x(t_0) + \int_{t_0}^{t_1}v[s_\theta](x_t,t)dt. \label{eq:odeslv}
}
The dual diffusion implicit bridge (DDIB) \cite{su2023dual} consists of two concatenated ODE solves for translating a source-domain sample $x_0^{(s)}$ to a target-domain sample $x_0^{(t)}$:
\iali{
    x_1^{(t)} =x_1^{(s)} &= \odeslv(x_0^{(s)}, s_\theta^{(s)}, 0, 1), \\
    x_0^{(t)} &= \odeslv(x_1^{(t)}, s_\theta^{(t)}, 1, 0).
}
Each ODESolve is carried out by an independently trained score on source (resp.~target) domain. The translation is cycle-consistent \cite{zhu2017unpaired} but \emph{unpaired}, since no explicit control via any predefined correspondences $(x_0^{(s)}, x_0^{(t)})$ is enforced. 

\subsection{Denoising Diffusion Bridge Models}
The denoising diffusion bridge model (DDBM) \cite{zhou2024denoising} considers the same diffusion process in \eqref{eq:f-sde}, but pinned to a fixed endpoint $y\in\bR^d$ almost surely via Doob’s $h$-transform:
\iali{
    dx_t &= \big[f(x_t,t)+g^2(t)\nabla_{x_t}\log q(x_1=y|x_t)\big]dt + g(t)dw_t.
}
A conditional generator $q(x_t|x_1=y)$ can be obtained from the reverse SDE: 
\iali{
    dx_t &= \big[f(x_t,t)-g^2(t)\big(\nabla_{x_t}\log q(x_t|x_1=y) 
    -\nabla_{x_t}\log q(x_1=y|x_t) \big)\big]dt + g(t)d\bar{w}_t. 
}
The endpoint $x_1$ is not restricted to follow a Gaussian prior, only the transition kernel $q(x_t|x_0,x_1)$ needs to be analytically tractable. DDBM trains the conditional score $s_\theta(x_t,t,x_1)$ on \emph{fully paired} data $(x_0,x_1)\sim q_{01}$ with the denoising bridge score matching loss \cite{zhou2024denoising}:
\iali{
    \min_\theta~ &\bE_{\substack{t\sim\cU(0,1),\,(x_0,x_1)\sim q_{01},\\x_t\sim q(x_t|x_0,x_1)}} 
    \Big[ \omega(t)\big\|s_\theta(x_t,t,x_1) - \nabla_{x_t}\log q(x_t|x_0,x_1) \big\|^2 \Big].
}

\subsection{Latent Diffusion Models}
The latent diffusion model (LDM) \cite{rombach2022high} enhances pixel-level diffusion models by incorporting an autoencoder over a latent space of reduced dimensionality. A pair of encoder $\cE_\phi$ and decoder $\cD_\psi$ are trained with a sum of a reconstruction loss, a perceptual loss, and a patch-based adversarial loss. In a follow-up stage, a diffusion network\footnote{The official LDM implementation follows DDPM's parameterization \cite{ho2020denoising} with a noise predictor $\epsilon_\theta$, which relates to the score function $s_\theta$ by $s_\theta(x_t,t)=\epsilon_\theta(x_t,t)/\sigma_t$. 
}
$s_\theta$ is trained on the latents of the original data samples $z=\cE_\phi(x)$.
For imagery data, a typical architectural choice for diffusion networks comes with U-Net \cite{RFB15}, coupled with cross-attention channels in the presence of conditioning \cite{rombach2022high}.

\section{Methodology}
\label{sec:method}

\subsection{Latent Aligned Diffusion Bridges}

\begin{algorithm}[t]
    \caption{LADM training}
    \begin{algorithmic}[1] \rm
        \Require paired source-target samples $(\hat{x}_{0,k}^{(s)}, \hat{x}_{0,k}^{(t)})_{k\in\cK}$, unpaired target-domain samples $(\tilde{x}_{0,l}^{(t)})_{l\in\cL}$, pretrained source-domain LDM $(\cE_{\phi'}^{(s)}, \cD_{\psi'}^{(s)}, s_{\theta'}^{(s)})$, pretrained target-domain autoencoder $(\cE_{\phi}^{(t)}, \cD_{\psi}^{(t)})$.
        \State Infer source-to-latent correspondences via \eqref{eq:s2l-infer}.
        \State Construct the coupling distribution $q_{01}^{(t)}(x_0^{(t)},x_1^{(t)})$ in \eqref{eq:mix-dist}.
        \State Train the LADM score $s_{\theta}^{(t)}$ via \eqref{eq:mix-sml}.
        \State \Return the LADM tuple $(\cE_{\phi}^{(t)}, \cD_{\psi}^{(t)}, s_{\theta}^{(t)})$.
    \end{algorithmic}
    \label{alg:train}
\end{algorithm}

\begin{algorithm}[t]
    \caption{\MethodName~sampling}
    \begin{algorithmic}[1] \rm
        \Require source-domain sample $x_0^{(s)}$, source-domain LDM $(\cE_{\phi'}^{(s)}, \cD_{\psi'}^{(s)}, s_{\theta'}^{(s)})$, target-domain LDAM $(\cE_{\phi}^{(t)}, \cD_{\psi}^{(t)}, s_{\theta}^{(t)})$.
        \State Perform source-to-latent translation:
        $x_1^{(t)} = \odeslv(\cE_{\phi'}^{(s)}(x_0^{(s)}), s_{\theta'}^{(s)}, 0, 1).$
        \State Perform latent-to-target translation:
        $x_0^{(t)} = \odeslv(x_1^{(t)}, s_{\theta}^{(t)}, 1, 0).$
        \State \Return the decoded target sample $\cD_\psi(x_0^{(t)})$.
    \end{algorithmic}
    \label{alg:infer}
\end{algorithm}

\paragraph{Partially paired correspondences}
Our task concerns translating a sample from the source domain $x_0^{(s)}$ to a sample in the target domain $x_0^{(t)}$, given paired source-target correspondences $(\hat{x}_{0,k}^{(s)}, \hat{x}_{0,k}^{(t)})_{k\in\cK}$ and unpaired target-domain data $(\tilde{x}_{0,l}^{(t)})_{l\in\cL}$. The paired correspondences are possibly sourced from neural predictors, human labelers, or a hybrid of both. Considering that high-quality labeling on cross-domain data are costly and scarce particularly if human efforts are involved, the partially paired setup is of high practical relevance.

\paragraph{Transferring correspondences on source domain}
If not available beforehand, a latent diffusion model (LDM) on the source domain need be trained following the standard pipeline from the LDM repository\footnote{\url{https://github.com/CompVis/latent-diffusion}}. Given the encoder-decoder $(\cE_{\phi'}^{(s)}$, $\cD_{\psi'}^{(s)})$ and the score $s_{\theta'}^{(s)}$ on the latent domain, one can infer a matched latent $\hat{x}_{1,k}^{(t)}$ from an encoded source sample $\cE_{\phi'}^{(s)}(\hat{x}_{0,k}^{(s)})$ via the ODESolve in \eqref{eq:odeslv}:
\iali{
    \hat{x}_{1,k}^{(t)} &= \hat{x}_{1,k}^{(s)} = \odeslv(\cE_{\phi'}^{(s)}(\hat{x}_{0,k}^{(s)}), s_{\theta'}^{(s)}, 0, 1). \label{eq:s2l-infer}
}
Thus, the original source-to-target correspondences $(\hat{x}_{0,k}^{(s)}, \hat{x}_{0,k}^{(t)})_{k\in\cK}$ are translated into paired latent-to-target correspondences $(\hat{x}_{1,k}^{(t)}, \hat{x}_{0,k}^{(t)})_{k\in\cK}$, provided that the source and target LDMs share a common latent space.

\paragraph{Semi-supervised learning on target domain}
With available latent-to-target correspondences in hand, we formulate the training of target-domain latent diffusion models as \emph{semi-supervised learning} \cite{KMRW14}. That is, we have both paired samples $(\hat{x}_{0,k}^{(t)}, \hat{x}_{1,k}^{(t)})_{k\in\cK}$ and unpaired samples $(\tilde{x}_{0,l}^{(t)})_{l\in\cL}$ as training data and, therefore, the coupling distribution $q_{01}^{(t)}(x_0^{(t)},x_1^{(t)})$ as a mixture distribution: 
\iali{
    & q_{01}^{(t)}(x_0^{(t)},x_1^{(t)}) = \frac{1}{|\cK|+|\cL|}\cdot 
    \Big( \sum_{k\in\cK}\delta_{(\hat{x}_{0,k}^{(t)}, \hat{x}_{1,k}^{(t)})}(x_0^{(t)},x_1^{(t)}) + \sum_{l\in\cL} \delta_{\tilde{x}_{0,l}^{(t)}}(x_0^{(t)})\otimes q_1(x_1^{(t)}) \Big). \label{eq:mix-dist}
}
With the (pretrained) autoencoder $(\cE_\phi^{(t)}$, $\cD_\psi^{(t)})$ on the target domain, the latent aligned diffusion model (LADM) is trained with the score matching loss sampled from the coupling distribution $q_{01}^{(t)}$:
\iali{
    \min_\theta & \bE_{\substack{t\sim\cU(0,1),\,(x_0^{(t)},x_1^{(t)})\sim q_{01}^{(t)},\\x_t=\alpha_t\cE_\phi^{(t)}(x_0^{(t)})+\sigma_t x_1^{(t)}}} 
    \Big[ \omega(t)\big\|s_\theta^{(t)}(x_t,t) - \nabla_{x_t}\log q(x_t|\cE_\phi^{(t)}(x_0^{(t)}))\big\|^2 \Big]. \label{eq:mix-sml}
}
The LADM training pipeline is depicted in Algorithm \ref{alg:train}.

\paragraph{Sampling with latent aligned diffusion bridges}
A source-domain LDM and a target-domain LADM together form a latent aligned diffusion bridge (\MethodName), depicted in Algorithm \ref{alg:infer}, for source-to-target translation. 
The source-to-latent (resp.~latent-to-target) translation is carried out by an ODESolve \eqref{eq:odeslv} involving the source LDM (resp.~the target LADM). On par with DDIB \cite{su2023dual}, cycle consistency is guaranteed in LADB (on the latent level).

\paragraph{Incorporating conditioning}
Our LADB framework is flexible to handle \emph{conditioning}, that is, each target sample $x^{(t)}_{0,i}$ is attached with a condition variable $\xi^{(t)}_i$ being a class label or a continuous embedding of some text caption. In the presence of conditioning, we parameterize the score network $s_\theta^{(t)}(x_t,t,\xi)$ with condition $\xi$, through either concatenation or cross-attention \cite{rombach2022high}. A class-conditioned \MethodName~for neural style transfer will be demonstrated in the supplementary.

\subsection{Extension to Multi-Source Translation}
The LADB framework admits a natural extension to \emph{multi-source domain translation}, depicted in Algorithms \ref{alg:train-ms} and \ref{alg:infer-ms}.
In the setup of multi-source LADM training, we are given paired and unpaired samples from $J$ source domains in total, i.e., $(\hat{x}_{0,k_j}^{(s_j)}, \hat{x}_{0,k_j}^{(t)})_{k_j\in\cK_j,\,j\in[J]}$ for paired and $(\tilde{x}_{0,l_j}^{(s_j)})_{l_j\in\cL_j,\,j\in[J]}$ for unpaired. The source domains can be as diversified as cross-modal \cite{socher2013zero}, but required to have pretrained LDMs $(\cE_{\phi'_j}^{(s_j)}, \cD_{\psi'_j}^{(s_j)}, s_{\theta'_j}^{(s_j)})_{j\in[J]}$ over a common latent domain.
For each source, the paired samples are encoded to their latent correspondences via
\iali{
    \hat{x}_{1,k_j}^{(t)} &= \hat{x}_{1,k_j}^{(s_j)} = \odeslv(\cE_{\phi'_j}^{(s_j)}(\hat{x}_{0,k_j}^{(s_j)}), s_{\theta'_j}^{(s_j)}, 0, 1). \label{eq:s2l-infer-ms}
}
The mixture coupling distribution $q_{01}^{(t)}(x_0^{(t)},x_1^{(t)})$ combines paired and unpaired samples from all sources:
\iali{
    q_{01}^{(t)}(x_0^{(t)},x_1^{(t)}) &= \frac{1}{\sum_{j=1}^J(|\cK_j|+|\cL_j|)}\cdot 
    \sum_{j=1}^J \Big(\sum_{k_j\in\cK_j}\delta_{(\hat{x}_{0,k_j}^{(t)}, \hat{x}_{1,k_j}^{(t)})}(x_0^{(t)},x_1^{(t)}) \notag\\&\qquad 
    + \sum_{l_j\in\cL_j} \delta_{\tilde{x}_{0,l_j}^{(t)}}(x_0^{(t)})\otimes q_1(x_1^{(t)}) \Big), \label{eq:mix-dist-ms}
}
from which a multi-source LADM can be trained with \eqref{eq:mix-sml}.

\begin{algorithm}[t]
    \caption{Multi-source LADM training}
    \begin{algorithmic}[1] \rm
        \Require paired samples $(\hat{x}_{0,k_j}^{(s_j)}, \hat{x}_{0,k_j}^{(t)})_{k_j\in\cK_j,\,j\in[J]}$, unpaired samples $(\tilde{x}_{0,l_j}^{(s_j)})_{l_j\in\cL_j,\,j\in[J]}$, pretrained source-domain LDMs $(\cE_{\phi'_j}^{(s_j)}, \cD_{\psi'_j}^{(s_j)}, s_{\theta'_j}^{(s_j)})_{j\in[J]}$, pretrained target-domain autoencoder $(\cE_{\phi}^{(t)}, \cD_{\psi}^{(t)})$.
        \State Infer source-to-latent correspondences via \eqref{eq:s2l-infer-ms}.
        \State Construct the coupling distribution $q_{01}^{(t)}(x_0^{(t)},x_1^{(t)})$ in \eqref{eq:mix-dist-ms}.
        \State Train the LADM score $s_{\theta}^{(t)}$ via \eqref{eq:mix-sml}.
        \State \Return the LADM tuple $(\cE_{\phi}^{(t)}, \cD_{\psi}^{(t)}, s_{\theta}^{(t)})$.
    \end{algorithmic}
    \label{alg:train-ms}
\end{algorithm}

\begin{algorithm}[t]
    \caption{Multi-source \MethodName~sampling}
    \begin{algorithmic}[1] \rm
        \Require weighted source-domain samples $(x_{0,i}^{(s_i)}, \rho_{i})_{i\in\cI}$, source-domain LDMs $(\cE_{\phi'_i}^{(s_i)}, \cD_{\psi'_i}^{(s_i)}, s_{\theta'_i}^{(s_i)})_{i\in\cI}$, target-domain LADM $(\cE_{\phi}^{(t)}, \cD_{\psi}^{(t)}, s_{\theta}^{(t)})$.
        \Ensure $\cI$ is a multiset supported on $(s_j)_{j\in[J]}$; $\rho_i>0~\forall i\in\cI$ and $\sum_{i\in\cI}\rho_i=1$.
        \For{$i \in \cI$}
        \State Perform source-to-latent translation:
        $x_{1,i}^{(s_i)} = \odeslv(\cE_{\phi'_i}^{(_i)}(x_0^{(s_i)}), s_{\theta'_i}^{(s_i)}, 0, 1).$
        \EndFor
        \State Compute a weighted average of latent correspondences:
        $x_1^{(t)} = \sum_{i\in\cI}\rho_ix_{1,i}^{(s_i)}.$
        \State Perform latent-to-target translation:
        $x_0^{(t)} = \odeslv(x_1^{(t)}, s_{\theta}^{(t)}, 1, 0).$
        \State \Return the decoded target sample $\cD_\psi(x_0^{(t)})$.
    \end{algorithmic}
    \label{alg:infer-ms}
\end{algorithm}

Sampling with multi-source LADB is highly versatile, compared to fully paired methods \cite{zhou2024denoising} and conditional models \cite{rombach2022high,zhang2023adding,zhao2023uni}. The user is free to translate a single source sample or an arbitrarily weighted combination of multiple source samples. In particular, the underlying weighted average among multiple sources is the Fr\'echet mean \cite{frechet1948elements} over the shared latent space in the same spirit of latent interpolation \cite{KiWe13,RMC16}.

\section{Related Work}
\label{sec:related}

Classic denoising diffusion and score matching methods \cite{ho2020denoising,song2021score,de2021diffusion} achieve an expressive generative model by gradually transforming a Gaussian noise to a data sample through a PF-ODE or an SDE. 
A series of unpaired translation methods, which were rooted in the Schr\"odinger bridge problem and dynamical optimal transport \cite{Leo14}, have recently been proposed for transferring data between arbitrary pairs of empirical distributions, including rectified flow \cite{liu2023flow}, flow matching \cite{lipman2023flow,tong2023improving}, bridge matching \cite{peluchetti2023nondenoising,shi2023diffusion}, and stochastic interpolants \cite{albergo2022building,albergo2023stochastic}. 
When the empirical distributions arise with known couplings (typical in inverse problems such as image restoration \cite{kawar2022denoising}), paired translation methods are better suited, among which bridge methods \cite{liu2023i2sb,li2023bbdm,zhou2024denoising} are prominent examples of using pinned diffusion processes to incorporate data couplings. 
While being more general-purpose, conditional diffusion \cite{rombach2022high,zhang2023adding,zhao2023uni} or conditional flow matching \cite{lipman2023flow} counts as another viable option for paired translation.

It is well understood that score/flow/bridge matching methods typically preserve time-dependent marginals but \emph{not} the coupling distribution of endpoints \cite{liu2022rectified,lipman2023flow,albergo2024stochastic}. 
Our proposed LADB framework is strongly inspired by rectification \cite{liu2023flow} and iterative Markovian projection \cite{peluchetti2023nondenoising,shi2023diffusion}. 
Different from these prior works which directly rectify correspondences between source and target distributions, in this work we leverage partially available source-target pairs to circuitously forge latent-target correspondences through pre-trained source-domain LDMs.

\section{Experiments}
\label{sec:results}

\subsection{Experimental Setup}
\paragraph{Datasets} We evaluate our method using the LSUN-Bedroom and LSUN-Churches datasets \cite{yu2016lsunconstructionlargescaleimage}. 
For depth-to-image translation and multi-source-to-image translation, we utilize depth images and segmentation masks from the LSUN-Bedroom dataset. 
Specifically, we randomly select 100,000 training images and obtain their paired correspondences using the open-source annotators: LeRes \cite{Wei2021CVPR} and DDPM-segmentation \cite{baranchuk2021labelefficient}. 

\paragraph{Task coverage}
We validate the proposed LADB through three experimental paradigms. 
First, we testify LADB in Algorithms \ref{alg:train} and \ref{alg:infer} on depth-to-image translation in Section \ref{sec:depth2img},
demonstrating its robustness to partially paired settings while balancing translation quality and fidelity.
Building upon this, in Section \ref{sec:multisrc} we instantiate Algorithms \ref{alg:train-ms} and \ref{alg:infer-ms} to image translation from two source modalities, 
namely depth maps and semantic segmentation masks.
Finally, we showcase LADB’s extensibility to multi-target translation with class-conditioned style transfer task (results in supplementary).
These tasks collectively demonstrate LADB’s core strengths: (1) semi-supervised latent alignment for data-efficient translation; 
(2) unified latent modeling for multi-source interpolation; (3) extensible conditioning mechanisms for multi-target generation.
Additionally, we apply the same workflow of \ref{sec:depth2img} to HED-to-image, canny-to-image, and sketch-to-image translation and refer to the supplementary for exemplary results.

\paragraph{Evaluation methods}
We compare our method with representative state-of-the-art approaches: (unpaired) DDIB \cite{su2023dual}, (fully paired) DDBM \cite{zhou2024denoising}, conditional LDM (CondLDM) \cite{rombach2022high}, ControlNet \cite{zhang2023adding} and (multi-conditioned) Uni-ControlNet \cite{zhao2023uni}.
For quantitative evaluation, we use a validation dataset of 1,000 samples at resolution $256 \times 256$ from LSUN-Bedroom, containing paired correspondences for translating from depth images to RGB images and from segmentation masks to RGB images.
We evaluate translation quality using Fréchet Inception Distance (FID) \cite{heusel2018ganstrainedtimescaleupdate} and Inception Scores (IS) \cite{barratt2018noteinceptionscore}, 
and measure perceptual similarity and translation faithfulness using LPIPS \cite{zhang2018perceptual} and MSE (in $[0, 1]$ scale).

\subsection{Depth-to-Image Translation}
\label{sec:depth2img}

\begin{table}[t]
  \renewcommand{\tabcolsep}{4pt}
  \centering
  \footnotesize
  \begin{tabular}{l l | r r c r}
  \toprule
  Method & \#Paired & FID $\downarrow$ & IS $\uparrow$ & LPIPS $\downarrow$ & MSE $\downarrow$ \\
  \midrule
  DDIB (pixel)            & 0  & 309.55 & 3.65 & 0.7491 & 0.3104 \\
  DDIB (latent)           & 0  & 123.95 & 3.86 &  0.7331 & 0.2237 \\
  \midrule
  DDBM (pixel)              & 50\% & 51.96 & 2.05 & 0.6320 & 0.1223 \\
                          & 100\% & 46.94 & 2.26 & 0.6348 & 0.1705 \\
  DDBM (latent)         & 10\% & 39.39 & 2.15 & {\color{teal}0.6145} & {\color{teal}0.1147} \\
                        & 25\% & 35.69 & 2.21 & 0.6585 &  {\color{teal}0.1117} \\
                        & 50\% &  {\color{blue}34.23}  & 2.23  & {\color{purple}0.5947}  &  {\color{purple}0.1118}  \\
                        & 100\% & 33.72 & 2.38 & 0.6030  & 0.1206 \\
  \midrule 
  CondLDM              &  50\% & 52.51 &  {\color{blue}2.59}  &  0.7728  & 0.1372  \\
                        & 100\%  & 44.65  &  2.72  & 0.6384  & 0.1303  \\
  ControlNet          &  50\%  & 43.05 &  {\color{purple}2.66}  &  0.7327  & 0.1709  \\
                        &  100\%  & 41.50 &  2.46  &  0.6660  & 0.1614  \\
  \midrule
  LADB (ours)           & 10\% & {\color{teal}33.29} & {\color{teal}2.23}  &  0.6499  & 0.1191 \\  
                        & 25\%  & {\color{teal}33.44}  &  {\color{teal}2.29}  &  {\color{teal}0.6375} &   0.1195 \\
                        &  50\% &  {\color{purple}33.78} & 2.35  & {\color{blue}0.6335} & {\color{blue}0.1125} \\
                        & 100\% & 34.78 & 2.43  & 0.6387 & 0.1052 \\
  \bottomrule
  \end{tabular}
  \caption{Depth-to-Image Translation: 
  (pixel) and (latent) stand for pixel- and latent-space variants. 
  The best (resp.~runner-up) metrics under 50\% paired are marked {\color{purple}red} (resp.~{\color{blue}blue}).
  Best metrics under 10\%/25\% paired are marked {\color{teal}green}.}
  \label{tab:benchmark}
  \vspace{-.8cm}
\end{table}

\paragraph{Setup} 
We evaluate our method against DDIB \cite{su2023dual}, DDBM \cite{zhou2024denoising}, Conditional LDM (CondLDM) \cite{rombach2022high} and ControlNet \cite{zhang2023adding} under varying percentages of paired correspondences during training.
Both pixel-space and latent-space variants of DDIB and DDBM are implemented for fair comparisons.
Conditional LDM is adapted by concatenating depth as a control signal.
We conduct thorough comparisons with DDBM (latent) across multiple partially paired settings (10\%, 25\%, 50\%), as it performs the closest to our method. 
Additionally, we experiment with a DDBM (latent) variant augmented by randomly pairing unpaired images from the full dataset with depth images (results in supplementary).
Further details are provided in the supplementary materials.

\begin{figure*}[t]
  \centering
   \includegraphics[width=\linewidth]{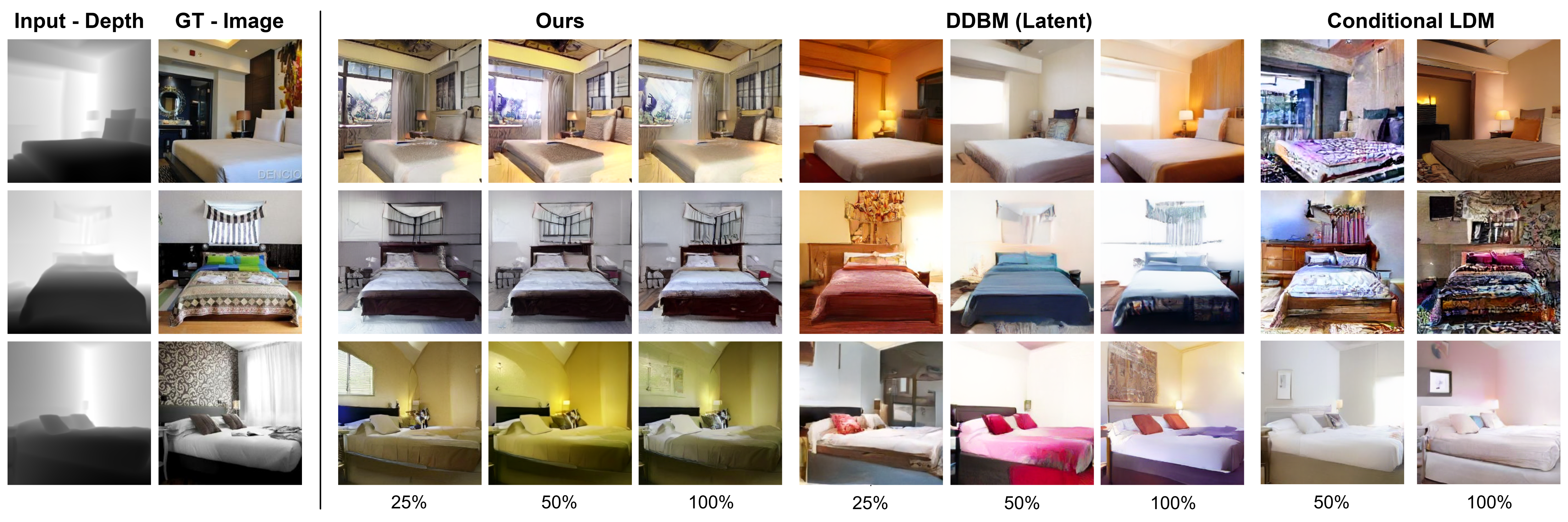}
   \caption{Qualitative comparisons. Our method produces translations that are visually coherent to the source. 
   It preserves fine-grained details (e.g., sheets and curtain) and avoids artifacts observed in CondLDM and DDBM.}
   \label{fig:benchmark_img}
  \vspace{-.5cm}
\end{figure*}

\paragraph{Results} 
As shown in Tab.~\ref{tab:benchmark}, our method achieves superior performances in generation quality (FID, IS) under partially paired settings, and is on par with baselines in translation fidelity (LPIPS, MSE).
Notably, LADB suffers less performance degradation compared to other baselines, as the percentage of paired data drops from 100\% to 10\%. 
Qualitative results in Fig.~\ref{fig:benchmark_img} further illustrate our method’s ability to preserve fine-grained details.

DDBM (latent) achieves high translation faithfulness (LPIPS, MSE) but lacks in generation quality (FID, IS). 
Conditional LDM and ControlNet prioritizes diversity (IS) at the cost of generation quality and fidelity (FID, LIPIS, MSE), while
DDIB falls behind in all metrics except for IS.

\paragraph{Discussion}
Our method’s success stems from its semi-supervised latent alignment, which learns to encapsulate domain-invariant latent features consistent to both paired and unpaired samples.
DDBM’s reliance on fully paired training causes overfitting, degrading fidelity (LPIPS, MSE) as the amount of paired data increases.
Conditional LDM’s dependence on dense supervision leads to catastrophic failure when dealing with partial pairing (FID, IS), as it cannot extrapolate to unseen configurations. 
In addition, Conditional LDM and ControlNet struggles to achieve high-quality and faithful translations (FID, LPIPS).
Furthermore, DDIB fails to produce coherent translations, likely due to significant distributional shift between the source and target domains.

Besides, it is observed that latent-space methods, such as our approach, DDBM (latent), Conditional LDM and ControlNet, outperform pixel-space counterparts. 
This is attributed to their superior ability to capture essential features and translate them faithfully across domains.

The results suggest that LADB behaves highly robust to lack of paired data, in a way to strike a balance between diversity and fidelity and preserve cross-domain consistency through latent-space alignment. This makes LADB a solution particularly competitive in data scarce translation tasks.

\subsection{Multi-Source-to-Image Translation}
\label{sec:multisrc}
\paragraph{Setup} We evaluate our method against UniControlNet (UniCtrl) \cite{zhao2023uni} and DDBM \cite{zhou2024denoising} for multi-source-to-image translation under the 25\% partially paired setting. 
We implement DDBM in latent space and, similar to our LADB, combine paired latent-target samples inferred from all source domains during training.
We utilize UniControlNet’s local adapter network to condition a pretrained LDM image diffusion model on both segmentation and depth inputs. 
We evaluate all methods under two settings: single-source input (depths or segmentations) and multi-source input $((\text{depth}, 0.5), (\text{segm}, 0.5))$. Note that UniControlNet only accepts both sources as input but not an interpolant of the two.
For more implementation details we refer to the supplementary materials.

\begin{table}[t]
  \renewcommand{\tabcolsep}{4pt}
  \centering
  \footnotesize
  \begin{tabular}{l | c c c c c c c c c}
  \toprule
  & \multicolumn{3}{c}{DDBM} & \multicolumn{3}{c}{UniCtrl} & \multicolumn{3}{c}{LADB} \\
  \cmidrule(lr){2-4} \cmidrule(lr){5-7} \cmidrule(lr){8-10}
  Metric & depth & segm & interp & depth & segm & both & depth & segm & interp \\
  \midrule
  FID $\downarrow$ & 36.03 & 38.65 & 55.43 & 41.55 & 41.96 & 37.03 & 34.10 & 38.67 & \textbf{34.72} \\
  IS $\uparrow$ & 2.21 & 2.35 & 2.66 & 3.02 & 3.06 & \textbf{2.92} & 2.30 & 2.33 & 2.38 \\
  LPIPS $\downarrow$ & 0.6181 & 0.5923 & \textbf{0.6403} & 0.6699 & 0.6728 & 0.6478 & 0.6494 & 0.6689 & 0.6798 \\
  MSE $\downarrow$ & 0.1108 & 0.0885 & 0.1451 & 0.1399 & 0.1360 & 0.1350 & 0.1191 & 0.1080 & \textbf{0.1275} \\
  \bottomrule
  \end{tabular}
  \caption{Multi-Source Translation results: 
  Scores are grouped by method and input source. 
  The best score for each metric in the multi-source setting is bolded.}
  \label{tab:multisource_benchmark}
  \vspace{-.5cm}
\end{table}

\begin{figure}[ht]
  \centering
  \includegraphics[width=0.8\linewidth]{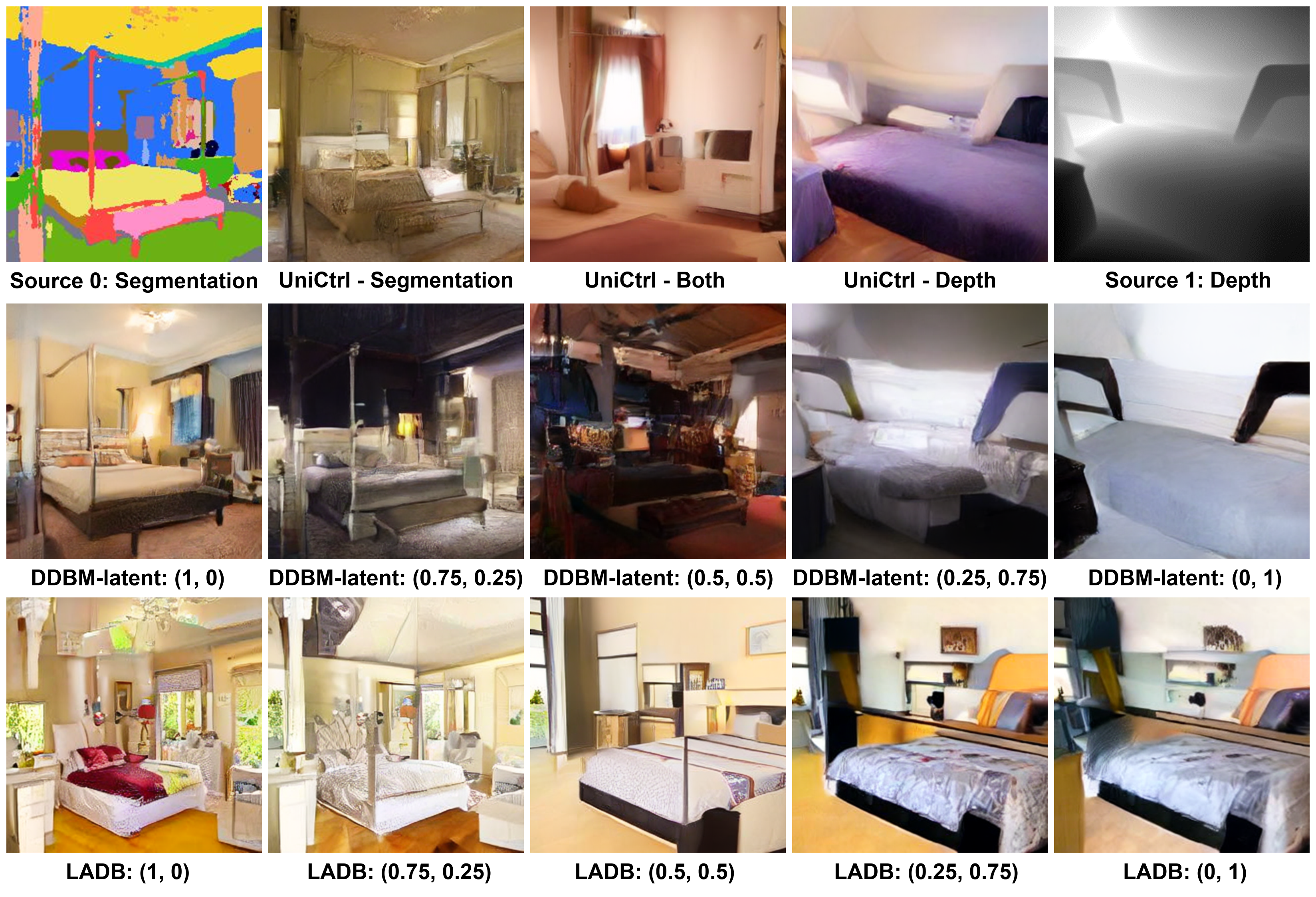}
  \caption{Multi-Source Translation visualization. Source 0 is a segmentation mask and Source 1 a depth map, with differing GT targets. Interpolated multi-source translations by UniControlNet, DDBM (latent), and LADB are displayed.}
  \label{fig:multisource_interpolation}
  \vspace{-.5cm}
\end{figure}

\paragraph{Results}  
The quantitative results in Tab.~\ref{tab:multisource_benchmark} show our method outperforms both baselines given multi-source inputs (FID: 34.72). 
Under single-source settings, LADB achieves superior fidelity over UniControlNet (FID, SSIM) and better generalization than DDBM (FID), 
in spite of DDBM’s marginally higher SSIM.

The qualitative results in Fig.~\ref{fig:multisource_interpolation} highlight LADB’s flexibility. 
Interpolation between depth- and segmentation-derived latents yields coherent blends of styles (such as illumination and colorization) and contents (such layouts and textures), 
a capability absent in rigidly conditioned baselines. 
In contrast, UniControlNet exhibits perceivable artifacts when plugged in with multiple modalities, and DDBM produces inconsistent structures across the interpolants.

\paragraph{Discussion} 
Our unified latent space enables adaptive fusion of cross-modal priors, addressing two key limitations of existing methods: 
UniControlNet’s reliance on fixed per-modality adapters and DDBM’s brittle pixel-level alignment. 

While UniControlNet improves with multi-source inputs, it remains inferior to LADB due to its inability to interpolate modalities or generalize to unseen combinations.
DDBM suffers larger performance drops during interpolation, as its strict pairing fails to generalize across interpolation between multiple sources.
In contrast, our semi-supervised alignment demonstrates highly smooth transitions in generation along interpolation (FID, SSIM),
thanks to distribution-level couplings through an aligned latent space.
The slight SSIM gap compared with DDBM (latent) in single-source segmentation translation may arise from inherent ambiguities in mask-to-image translation,
where sparse semantic cues challenge latent alignment. 

By projecting source data onto a unified latent space, our framework generalizes to unseen combinations without retraining (Fig.~\ref{fig:multisource_interpolation}), offering a scalable solution for real-world domain translation with incomplete or heterogeneous data.

%
%
%
%
\bibliographystyle{splncs04}
\bibliography{094-main}

\end{document}